\documentclass[a4paper]{article}

\usepackage{INTERSPEECH2021}
\usepackage{lipsum}
\usepackage{url}
\usepackage[inline]{enumitem}
\usepackage{balance}
\usepackage[absolute,overlay]{textpos}

\title{Speech Detection For Child-Clinician Conversations In Danish \\For Low-Resource In-The-Wild Conditions: A Case Study}
\name{Sneha Das$^1$, Nicole Nadine Lønfeldt$^2$, Anne Katrine Pagsberg$^{2, 3}$, Line. H. Clemmensen$^1$}
\address{
  $^1$Department of Applied Mathematics and Computer Science, Technical University of Denmark\\
  $^2$Child and Adolescent Mental Health Center, Copenhagen University Hospital, Capital Region\\
  $^3$Faculty of Health, Department of Clinical Medicine, Copenhagen University}
\email{sned@dtu.dk}

\begin{document}

\maketitle
\begin{textblock*}{5cm}(2cm,3cm) 
{\it Submitted to Interspeech 2022}
\end{textblock*}

\begin{abstract}
Use of speech models for automatic speech processing tasks can improve efficiency in the screening, analysis, diagnosis and treatment in medicine and psychiatry. However, the performance of pre-processing speech tasks like segmentation and diarization can drop considerably on in-the-wild clinical data, specifically when the target dataset comprises of atypical speech. In this paper we study the performance of a pre-trained speech model on a dataset comprising of child-clinician conversations in Danish with respect to the classification threshold. Since we do not have access to sufficient labelled data, we propose few-instance threshold adaptation, wherein we employ the first minutes of the speech conversation to obtain the optimum classification threshold. Through our work in this paper, we learned that the model with default classification threshold performs worse on children from the patient group. Furthermore, the error rates of the model is directly correlated to the severity of diagnosis in the patients. Lastly, our study on few-instance adaptation shows that three-minutes of clinician-child conversation is sufficient to obtain the optimum classification threshold. 
\end{abstract}

\section{Introduction}

Recent advancements in speech processing and the successful application of deep-learning for speech modelling tasks have aided in the emergence of speech interfaces. Speech signals are used in wide applications, for instance in the educational sector to gauge student learning, in commercial sectors for job placements, in the medical domain for the transcription of medical records and in psychiatry to help psychologist in delivering therapy and aiding in diagnosis. These applications employ a multitude of speech processing tasks, like automatic speech recognition, speech emotion recognition, speech detection, speaker diarization, etc.

Speech signals in real-world applications vary greatly from the signals the speech models are trained on. The sources of variations could be gender, age, noise and consistency in the recording conditions, languages, number of speakers, and noise and consistency in the recording conditions. Therefore, prior to applying the state-of-the-art speech models on the target dataset, speech and audio signals will need to be pre-processed. If the speech samples are in the form of a conversation, the signals will first have to be segmented into speech and non-speech regions. Subsequently, each detected speech segment must be assigned to a specific speaker, a process termed speaker diarization. Current state-of-the-art methods approach these steps in a modular manner where the segmentation and diarization are different steps, or use an end-to-end approach. In a modular setup, the accuracy of the diarization step depends on the accuracy of segmentation. Therefore, the accuracy of the segmentation task is critical in for the performance of the downstream speech processing tasks in such systems.

Models for pre-processing speech can be supervised, unsupervised or a combination of the two. Fully supervised methods will work when there are large amounts of labelled data. Labelling all available data is not always possible due to the lack of resources, in terms of time, money, and people. When limited labelled data is available, using a pre-trained model and fine-tuned with the available data is a viable alternative. However, the challenge here is that the data must be relatively uniform in terms of recording conditions, speakers, and noise levels in order for a model to yield reasonable performance. These ideal conditions are rarely available in-the-wild.

Within the scope of this paper, we explore pre-processing quality in terms of speech detection in conversations between a clinician and a child. The conversations are in Danish and between adults and children between ages 8-17, implying a wide range of voice types. Furthermore, since the application is clinical data for psychiatric disorders, the systems should be equally good with non-typical speech. In addition, the audio obtained is in-the-wild implying that the audio quality is not regulated. In other words, the audio conversations comprise of large variations in terms of speech and noise levels, the microphone on the video recorder is solely used to record whereby there may be considerable channel noise in the audio recordings. Under these conditions, the above listed approaches do not provide acceptable performance as required by our application.

Studies on the use of speech processing in medicine have mainly focused on medical record summarization and speech recognition for medical conversations\cite{chiu2017speech, 47001, enarvi2020generating, soltau2021understanding}. The authors in~\cite{47001} describe their experience in developing an automatic speech recognition to transcribe medical conversations. Their study was on English records and has access to sufficient labelled data to train models for their application. Further on, methods have been proposed for speech-to-text on medical conversations and transcribing medical conversations with high accuracy~\cite{enarvi2020generating, soltau2021understanding}. However, there  is a need for studies on the quality of pre-trained speech processing models for speaker groups in in-the-wild clinical conversations.

In this work, we use a state-of-the-art pre-trained model for speech segmentation~\cite{ravanelli2021speechbrain}. The pre-trained model has a default threshold in classifying segments to speech and non-speech regions, optimized over the training dataset. Ideally, this threshold should be fine-tuned to the target dataset. However, fine-tuning the threshold requires sufficient labelled data, to which we do not have access. Also, a global threshold for the target dataset will work when the variation between samples is small. Therefore, the goal in this work is to obtain the optimum threshold for each clinician-child pair with least amount of labelled data. We hope that with this work, we can employ automatic speech segmentation and leverage human labels together to efficiently segment data with improved accuracy on our clinical dataset. The contributions of this work are the following: \begin{enumerate*}\item We investigate the performance of the pre-trained model in its default setting, and analyse differences in its performance between the different speakers and the participant groups. \item We show that the performance of the segmentation algorithm is directly correlated to OCD-severity scores. \item We explore few-instance threshold adaptation, wherein we employ the first few minutes to adapt the threshold. The analysis questions are: (a)~how much labelled data is {\it sufficient} to improve segmentation quality for each clinician-child pair? (b)~what is the limit on performance improvement by modifying the classification threshold. To the best of our knowledge, this is among the first works that investigates the performance of pre-processing speech models on the different speaker groups in in-the-wild clinical conversations.  
\end{enumerate*}

    \section{Resource Description}
    \subsection{Dataset}
    \label{Sec:dataset}
The audio conversations used in this study is from the video recordings of clinicians interviewing children (8-17 years) with the Kiddie Schedule for Affective Disorders and Schizophrenia (K-SADS)~\cite{puig1986kiddie}. The K-SADS is a semi-structured interview designed to establish psychiatric diagnoses in child between 6 and 18 years of age. The interview generally lasts 30 to 90 minutes. We used audio from a randomized clinical trial and case-control study of obsessive-compulsive disorder (OCD), the TECTO trial~\cite{pagsberg2022family} . Thus, the participants in this cohort included 8-17 year old children with OCD and children with no psychiatric diagnoses. Ten minutes of the interviews from each clinician-child pair were used due to the limited availability of labelled data. The video was recorded using an Sony video camera in an arbitrary position in different rooms. The on-camera microphone was employed to record the conversation. Therefore, the audio quality is variable. We included 5 participants who were diagnosed with OCD and 5 control participants. Thus, the dataset employed in this study comprises of multiple sources of variation, due to noise from differences in recording scenarios and non-typical speech due to population differences~(age, sex, diagnostic status).

Clinical severity ratings are assigned to participants diagnosed with OCD using the Children’s Yale-Brown Obsessive
Compulsive Scale (CY-BOCS) ~\cite{scahill1997children}. The CY-BOCS is a semi-structured interview in which clinicians rate 10 items on the severity of obsessions and compulsions over the past week from 0 to 4. An obsession severity score is calculated by summing the first five items (obsession severity rating range: 0-20). A compulsion severity is obtained by summing items 6-10 (compulsion severity range:0-20). A total OCD severity rating is computed by summing all 20 items and ranges between 0-40. To be included in the trial, participants required a severity rating of at least 16 at baseline. Severity scores from 16-23 are considered moderate and scores from 24 to 40 are deemed severe to extreme. The control patients were not interviewed with the CY-BOCS and so do not have a CY-BOCS score.
    \subsection{Pre-trained model}
We used the vad-crdnn-libriparty pre-trained model from speech brain for speech detection~\cite{ravanelli2021speechbrain} due to its flexibility and modular structure, which is a favourable quality when applying models to clinical data due to the tractability of errors. The pre-trained model is based on CRDNN which is a model that combines the convolution neural network~(CNN), recurrent neural network~(RNN) and fully connected network. The model accepts audio segments as input and yields a posterior probability frame-level/segment level posterior probability as output. Finally, a threshold is applied on the output posterior probability to classify the segments as speech and non-speech. In this work, we modify this classification threshold in the final classification step and evaluate the performance of the models on the considered data. 

    \subsection{Target device specification}
The target device used in this work comprised of an Intel(R) Core(TM) i5-6300U CPU @ 2.40 GHz, 7.99~GB RAM and runs Windows~10 operating system. Due to the unavailability of computation resources, the choice and design of the inference and fine-tuning algorithms are constrained for this study. 
\begin{figure}[!tbh]
\centering
\includegraphics[width=0.99\columnwidth]{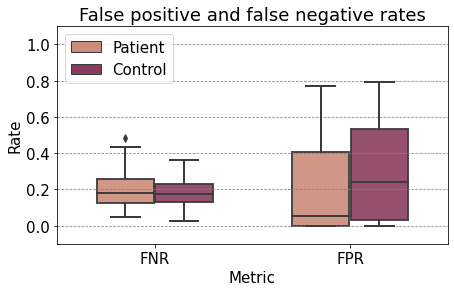}
\caption{False negative rate~(FNR) and false positive rate~(FPR) between the patient and control groups.}
\label{fig:FNR_FNR_PC}
\end{figure}
    \section{Pre-trained Model Performance}
    We investigate the performance of the model in its default threshold for speech non-speech on the underlying participant groups in the audio recordings.

\noindent
{\bf Patient versus Control groups: }
The false negative rate~(FNR) and the false positive rate~(FPR) are computed as follows:
\begin{equation}\label{eq:1}
R_{\text{FNR}}=\frac{n_{\text{P}_\text{NS}}}{n_{\text{T}_\text{S}}},  \quad R_{\text{FPR}}=\frac{n_{\text{P}_\text{S}}}{n_{\text{T}_\text{NS}}},  
\end{equation}
where $n_{\text{P}_\text{NS}}$ indicates the time samples incorrectly predicted as non-speech, $n_{\text{P}_\text{S}}$ indicates the time samples incorrectly predicted as speech, $n_{\text{T}_\text{NS}}$ is the total non-speech duration in the signal and $n_{\text{T}_\text{S}}$ shows the total speech duration in the signal. 
The above measures are computed over 10 one-minute long segments for each cliniciac-child recording. We do this to gain insights into the performance consistency of the individual speakers in a speaker pair. The distribution of the FNR and FPR or the patient and control groups are presented in Fig.~\ref{fig:FNR_FNR_PC}. For the FNRs between the participant groups, the medians are similar, but the inter-quartile range~(IQR) is larger for the patient group. This could be because of variations in the overall duration of clinician and child segments in each conversation. Although, the FPRs seem to be much higher than the FNRs, it remains consistent over all the experiments in this study, whereby we focus our analysis on FNRs for the remaining study.

\begin{figure}[!tbh]
  \centering
    \includegraphics[width=0.99\columnwidth]{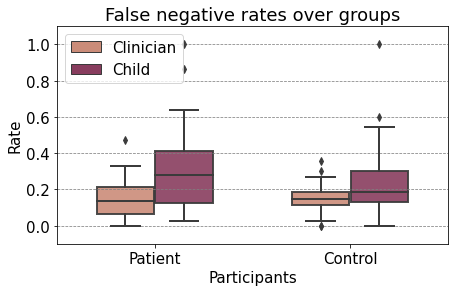}
  \caption{FNR of clinician and child duration in the recordings for the two participant groups.}
  \label{fig:default_eval_TvsC}
\end{figure}

\noindent
{\bf Clinician versus Child groups: }
The length of speech corresponding to clinician and child in an audio conversation could vary between each clinician-child pair, whereby the group differences between the groups could diminish. Therefore, to obtain balanced insights into the performance of the default model on the different participant groups, we computed the FNR following Eq.~\ref{eq:1} separately for the child and the clinician in each recording:
\begin{equation}
R_{\text{FNR}|P}=\frac{n_{\text{P}_\text{NS}|P}}{n_{\text{T}_\text{S}|P}}, \quad P\in\{Clinician, Child\}.
\end{equation}
From the results presented in Fig.~\ref{fig:default_eval_TvsC}, we observe that the FNR distribution for the clinician is similar over the patient and control groups. In contrast, the median and the IQR range for the FNR of the patient children is relatively higher than the control children. This indicates a relatively sub-optimal performance for the children in patient group with the default threshold. 

\begin{figure}[!tbh]
  \centering
   \includegraphics[width=0.99\columnwidth]{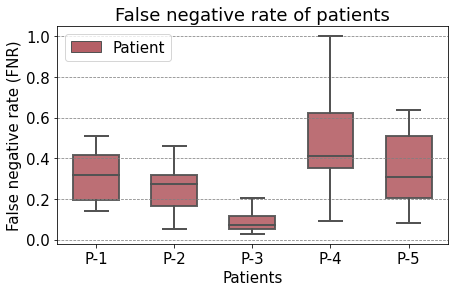}\\    \includegraphics[width=0.99\columnwidth]{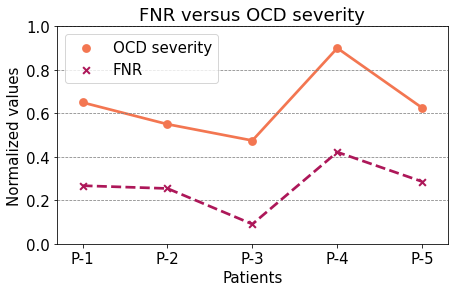}
  \caption{FNRs of children in patient group only. (Top)~Distribution of the FNRs, (Bottom)~OCD-severity versus False negative rates.}
  \label{fig:severe_FNR}
\end{figure}

\noindent
{\bf OCD-severity versus FNR: }
We explore the correspondence between the CY-BOCS severity scores described in Sec.~\ref{Sec:dataset} and the FNR of the patients. The results are shown in Fig.~\ref{fig:severe_FNR}, with box-plots showing the FNR distribution of the children in the patient group over the entire conversation, and the OCD-severity and FNRs plotted for each patient. Note that the OCD-severity has been normalized to match the FNR scaling. From the plots, we observe a one-to-one correspondence between the OCD-severity and the FNRs. The reason for this could be that OCD-severity is proportional to atypical voice characteristics. In other words, higher the OCD-severity, the underlying speech signal is further away from average speech characteristics. Since speech detection models learn average speech characteristics, atypical speech shows higher error rates. 

\section{Adapting Speech Threshold}
In this section, we present the FNR rates when the speech-non-speech threshold is adapted for each segment. Furthermore, since we do not have access to labels beyond the first minutes of the recordings, we investigate the length of speech sufficient to improve performance for each clinician-child pair. 

\begin{figure}[!tbh]
  \centering
    \includegraphics[width=0.66\columnwidth]{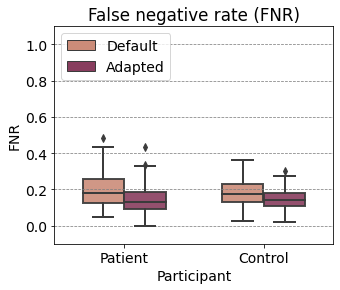}\includegraphics[width=0.33\columnwidth]{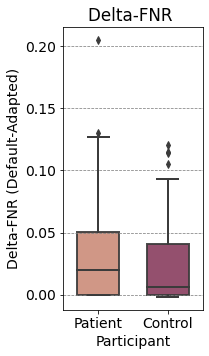}
  \caption{FNR rates after adapting thresholds for each one-minute long segment.}
  \label{fig:Problem_description}
\end{figure}

Fig.~\ref{fig:Problem_description} presents the optimum threshold that minimizes the FNR for each segment in a recording. The box-plots illustrate the absolute spread and the delta improvement in the FNR between (delta-adapted). We observe that improvement in the FNR for patients is higher than for the control group. However, optimizing the threshold for all segments is infeasible due to insufficient resources to label data. In the following section, we explore threshold adaptation using few instances of a clinician-child pair. 

\begin{figure}[!tbh]
  \centering
   \includegraphics[width=0.99\columnwidth]{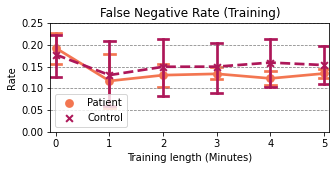}\\    \includegraphics[width=0.99\columnwidth]{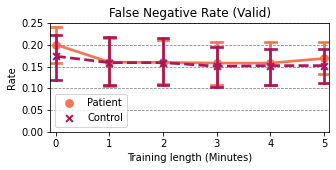}\\    \includegraphics[width=0.99\columnwidth]{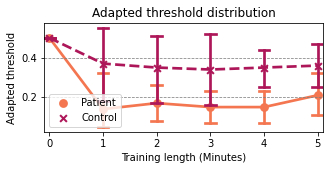}
  \caption{FNR rates for the training and test segments and distribution of the threshold for incremental segment lengths. (Top)~FNR between participant groups for training segments, (Center)~FNR for test segments, (Bottom)~Distribution of the adapted thresholds over different training segment length.}
  \label{fig:adapted_th}
    \vspace{-0.3cm}
\end{figure}

\noindent
{\bf Few-instance threshold adaptation: }To incrementally adapt the threshold, we compute the optimum threshold by minimizing the FNR and     FPR, 
\begin{equation}
    \arg \min_{th} \quad  \frac{\sum_{t=1}^{T}n_{\text{P}_\text{NS}}(t, th)}{\sum_{t=1}^{T}n_{\text{T}_\text{S}}(t, th)}+\frac{\sum_{t=1}^{T}n_{\text{P}_\text{S}}(t, th)}{\sum_{t=1}^{T}n_{\text{T}_\text{NS}}(t, th)}, 
\end{equation}
where $k$ is the segment index, and $th$ is the threshold being optimized. We optimize the threshold incrementally over all first five-minutes of the audio conversation and compute the FNR. The validation is done over the last five-minutes segments. The training and validation FNR, and the adapted threshold values over the iterations are presented using the mean and 95\% confidence intervals in Fig.~\ref{fig:adapted_th}. The goal is to have lower mean and variance in the FNRs and the variance of adapted threshold should be low as well. From the training and validation FNRs, we observe that there is a drop in the FNR already after one-minute of training. However, the variance of the FNR in the training dataset reduces as the length of the training increases as is anticipated. In terms of the adapted threshold, the size of the error bars are lowest at three-minutes training length for the patient group. Furthermore, the mean of the adapted threshold is much lower for the patient group, implying lower probability scores by the algorithm for speech regions in this group. 

\begin{figure}[!tbh]
  \centering
   \includegraphics[width=0.99\columnwidth]{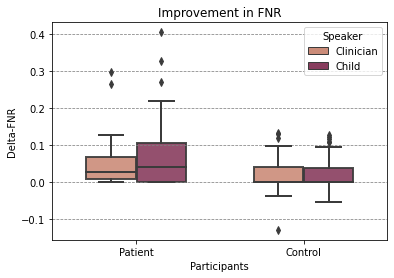}
  \caption{Improvement in FNR with few-instance threshold adaptation between the participant groups.}
  \label{fig:adapted_th_who}
  \vspace{-0.6cm}
\end{figure}

The range of improvement in FNRs using this threshold adaptation approach, relative to the default threshold, $th=0.5$ is illustrated in Fig.~\ref{fig:adapted_th_who}. We observe that the improvement for the control is similar over both the speakers. However, in the paient group, FNR improvement for child is higher than for the clinician, with a larger IQR. Furthermore, in the control group threshold adaptation results in reduction in performance with respect to the FNRs. In contrast, using threshold adaptation for the patient groups seems to improve performance of the segmentation algorithm for all samples and segments.

\section{Discussion and Conclusions}
Use of deep learning models for speech processing on clinical data in medicine and psychiatry can greatly accelerate screening, evaluation, diagnosis and treatment. However, before applying machine learning models to clinical data, it is essential to gauge the performance of these models on the target dataset. Furthermore, it is necessary that the models perform equally well on all participant groups~(patient, control) in the target dataset. This is rooted in the fact that deep learning models are often as good as the data it is trained on, and under conditions of domain mismatch the models perform poorly.
In this study we addressed the questions: \begin{enumerate*}
\item How does a state-of-the-art pre-trained model perform on an in-the-wild clinical dataset? \item Does the algorithm perform equally well on all the participant groups?
\item What amount of data is necessary to improve the performance by adapting the classification threshold? \item What is the limit on the performance gains with threshold adaptation. 
\end{enumerate*}
We learned that the model with the default classification threshold performs better on the control group children relative to the children from the patient group. In contrast, its performance is consistent for therapist from both participant groups. A critical finding from our study is that, the FNR rates seems to be perfectly correlated with the CY-BOCS OCD-severity scores. We therefore conclude that the model is sensitive to atypical characteristics of speech, and children with higher OCD-severity scores may have different speech characteristics to the speech characteristics of children from the control group. Furthermore, our experiments using few-instance threshold adaptation indicates that three-minutes of labelled data for each clinician-child pair is sufficient to improve segmentation accuracy. The improvement is higher for children in the patient group, although the distribution of improvement with threshold adaptation has a high IQR for this group. 

From our experiments, we conclude that threshold adaptation can improve FNR rates by upto 5\%. Therefore, going forward we should consider fine-tuning the weights of the pre-trained model. While this will not only require a lot more labelled data, due to the high variation in the recordings between clinician-child data the fine-tuned model may not generalize well. Therefore, it may be worth exploring personalized pre-processing models for each pair, despite the overhead. 

\bibliographystyle{IEEEtran}
\nocite{*}
\bibliography{ref}
\balance
\end{document}